\documentclass[journal]{IEEEtran}
\usepackage{amsmath,amsfonts}
\usepackage{algorithmic}
\usepackage{algorithm}
\usepackage{array}
\usepackage[caption=false,font=normalsize,labelfont=sf,textfont=sf]{subfig}
\usepackage{textcomp}
\usepackage{stfloats}
\usepackage{url}
\usepackage{verbatim}
\usepackage{graphicx}
\usepackage{cite}
\usepackage{graphicx}
\usepackage{amsmath}
\usepackage{amssymb}
\usepackage{booktabs}
\usepackage{enumitem}
\usepackage{multicol}
\usepackage{multirow}
\usepackage{amsthm}
\usepackage{mathrsfs}
\usepackage{diagbox}
\usepackage{bm}
\usepackage{newfloat}
\usepackage{xspace}
\usepackage{xcolor}
\usepackage{colortbl}
\usepackage{nicematrix}
\usepackage{stfloats}
\usepackage{verbatim}
\usepackage{cite}
\usepackage{tikz}
\usepackage{comment}
\usepackage{amsmath,amssymb} 
\usepackage{color}
\usepackage{lipsum}
\usepackage{amsmath}
\usepackage{wasysym}
\usepackage{graphics} 
\usepackage{epsfig} 
\usepackage{makecell}
\usepackage{multirow}
\usepackage{mwe,lipsum}
\usepackage{array,multirow}
\usepackage{float}
\usepackage{comment}
\usepackage{pifont}
\usepackage{tabularx}
\usepackage{arydshln}

\begin{document}

\title{Depth-guided Texture Diffusion for \\Image Semantic Segmentation}

\author{Wei Sun, Yuan Li, Qixiang Ye,~\IEEEmembership{Senior Member,~IEEE,} Jianbin Jiao,~\IEEEmembership{Member,~IEEE,} Yanzhao Zhou
\thanks{Wei Sun, Qixiang Ye and Yanzhao Zhou are with the School of Electronic, Electrical, and Communication Engineering,
 University of Chinese Academic of Sciences (UCAS), Beijing 101408, China. Yanzhao Zhou is corresponding author. e-mail: ({sunwei162@mails.ucas.ac.cn; qxye@ucas.ac.cn; zhouyanzhao@ucas.ac.cn)}}

\thanks{Yuan Li and Jianbin Jiao is with the School of Emergency Management Science and Engineering, University of Chinese Academic of Sciences (UCAS), Beijing 101408, China. e-mail: ({liyuan23@ucas.ac.cn; jiaojb@ucas.ac.cn)}}
}



\maketitle

\begin{abstract}
Depth information provides valuable insights into the 3D structure especially the outline of objects, which can be utilized to improve the semantic segmentation tasks. However, a naive fusion of depth information can disrupt feature and compromise accuracy due to the modality gap between the depth and the vision.
In this work, we introduce a Depth-guided Texture Diffusion approach that effectively tackles the outlined challenge. Our method extracts low-level features from edges and textures to create a texture image. This image is then selectively diffused across the depth map, enhancing structural information vital for precisely extracting object outlines. By integrating this enriched depth map with the original RGB image into a joint feature embedding, our method effectively bridges the disparity between the depth map and the image, enabling more accurate semantic segmentation.
We conduct comprehensive experiments across diverse, commonly-used datasets spanning a wide range of semantic segmentation tasks, including Camouflaged Object Detection (COD), Salient Object Detection (SOD), and indoor semantic segmentation. With source-free estimated depth or depth captured by depth cameras, our method consistently outperforms existing baselines and achieves new state-of-the-art results, demonstrating the effectiveness of our Depth-guided Texture Diffusion for image semantic segmentation.
\end{abstract}


\begin{figure}[t]
  \centering
  \includegraphics[width=\columnwidth]{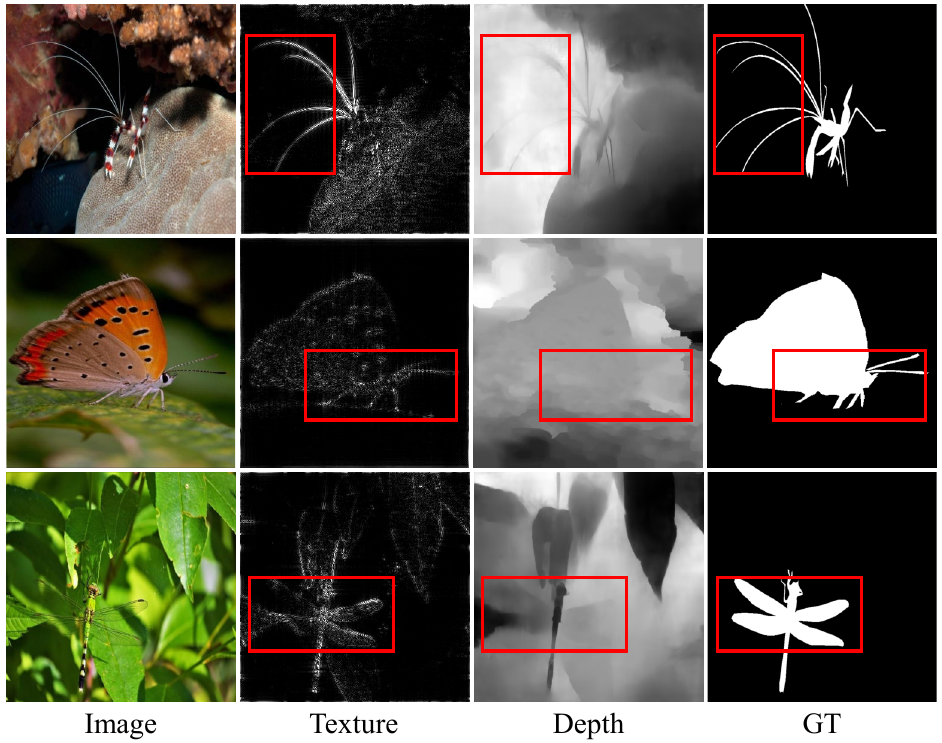} 
  \caption{Visual samples illustrating the absence of detailed textures in depth maps compared to the corresponding texture images and ground truth (GT) images.}
  \label{fig:teaser}
\end{figure}

\section{Introduction}
\IEEEPARstart{I}{mage} semantic segmentation is a crucial task in computer vision, designed to divide an image into regions that are meaningful based on visual characteristics. This process is vital for applications including object recognition, scene understanding, and image editing. Traditional approaches primarily utilize color, contour, and shape cues from 2D images to conduct semantic segmentation. However, these methods often fail to capture the 3D structure of the scene, which can result in less accurate outcomes, especially in complex settings.

To address this challenge, researchers have integrated depth information into the segmentation process, which offers valuable geometric context. Depth information enriches the understanding of the scene by revealing crucial cues, such as the distances of objects and their occlusion relationships, which significantly aid in the segmentation task. By incorporating depth into the input, segmentation outcomes become more accurate and robust.

Recently, there have been notable research efforts in RGB-D semantic segmentation tasks, which combines color and depth data for enhanced analysis. Researchers have developed fusion models that leverage both RGB image and depth map to better exploit these complementary modalities. In some studies, depth maps are integrated as an additional channel in the early stages of the model~\cite{he2017std2p, husain2016combining, wu2023source}. Similarly, fusion methods have been designed to combine features extracted from both RGB and Depth data~\cite{xing2019coupling, seichter2021efficient, park2017rdfnet, jiang2018rednet, hu2019acnet, hazirbas2017fusenet, fooladgar2019multi, chen2020bi}. However, directly fusing depth information with image features poses a challenge due to the intrinsic distribution gap between the two modalities. As shown in Fig.~\ref{fig:teaser}, depth maps are inherently different from traditional RGB images. Depth maps, irrespective of their origin from depth estimation models or depth cameras, are characterized by their lack of texture and fine details, a stark contrast to what is typically observed in RGB representations. The inherent limitations of depth-sensing technology and depth estimation algorithms mean that these maps are better at representing broader spatial relationships rather than capturing the intricate textures and details. Furthermore, by reducing three-dimensional space to one-dimensional pixel intensities indicating distances, depth maps fail to capture the textural and color nuances inherent to RGB images. This discrepancy can lead to semantic mismatches when depth and RGB features are combined using simplistic methods without considering their fundamental differences.

In this study, we introduce a novel method, depth-guided texture diffusion, tailored to enhance the compatibility between depth and 2D images by selectively accentuating textural details such as object outlines within depth maps. This technique infuses the depth map with texture-like cues, effectively bridging the gap between depth and vision modality. To ensure the integrity of object structures in the depth map after texture diffusion, we deploy a structural loss function. This function helps preserve structural consistency, thereby minimizing discrepancies between the depth and RGB images and reducing information loss during fusion. Furthermore, we perform an integrated encoding of the texture-refined depth and RGB image, creating a visual prompt that embeds texture-aware depth information into the original model. This strategy enhances the accuracy of object extraction and boosts overall semantic segmentation performance.

To evaluate the effectiveness of our proposed method, we conduct extensive experiments across a diverse range of datasets. For camouflaged object detection, we utilize source-free estimated depth, while for salient object detection and indoor semantic segmentation, we employ depth captured by depth cameras. The experiment results demonstrate that our depth-guided texture diffusion consistently and significantly outperforms baseline methods, demonstrating superior performance over state-of-the-art segmentation methods.

In summary, this study introduces a depth-guided texture diffusion approach that enhances image semantic segmentation. By bridging the gap between depth data and 2D image, our method not only enriches the depth information but also improves the segmentation accuracy. This is achieved by incorporating textural cues into depth maps, aligning them more effectively with RGB images and enhancing the model’s interpretation of complex scenes.
Extensive experimental results across various datasets for salient object detection, camouflaged object detection, and indoor semantic segmentation confirm the effectiveness of our approach. These results demonstrate our method's capability to meet and exceed existing benchmarks, indicating its potential to contribute to advancements in depth-guided image segmentation.

The main contributions of this paper are threefold:
\begin{itemize}
  \item We introduce a novel depth-guided texture diffusion approach that bridges the gap between depth and vision modality, enabling the effective utilization of 3D structural information to achieve more accurate semantic segmentation.
  
  \item We introduce structural consistency optimization to ensure that depth maps align with the structural integrity of RGB images after texture diffusion, thereby further boosting the model's performance and robustness.
  
  \item We establish new state-of-the-art benchmarks on multiple commonly used datasets covering the task of camouflaged object detection, salient object detection and indoor semantic segmentation. This underlines the versatility and effectiveness of our depth-guided texture diffusion in a wide range of real-world applications.
\end{itemize}

\section{Related Work}
\subsection{RGB-D Scene Parsing}
In the landscape of contemporary computer vision, three vibrant research areas stand out in the RGB-D scene parsing domain: indoor semantic segmentation, salient object detection, and camouflaged object detection. Indoor semantic segmentation specifically aims to assign class labels to each pixel in indoor scenes, effectively delineating areas such as furniture, walls, and appliances to enhance scene understanding and spatial analysis in complex indoor spaces. Salient object detection is geared towards highlighting elements that naturally stand out, demanding focus in a scene. On the other hand, camouflaged object detection endeavors to uncover objects that are designed to blend seamlessly with their surroundings, posing a unique set of challenges. 

The integration of RGB images with depth maps presents complex challenges due to the inherent disparities in their modal distributions. While RGB images capture detailed visual information, depth maps offer essential spatial relationships that are not immediately discernible in RGB data. This discrepancy often complicates the fusion process, requiring sophisticated approaches to fully leverage the complementary strengths of both modalities. Significant research efforts are directed towards developing innovative fusion techniques to effectively bridge this gap and enhance overall scene understanding.

Substantial strides have been made in these fields through concerted efforts in developing robust fusion modules that combine the fine-grained detail available in RGB imagery with the spatial context provided by depth data. Models like CMX (Zhang \textit{et al.})~\cite{zhang2023cmx}, TokenFusion (Wang \textit{et al.})~\cite{wang2022multimodal}, and HiDANet (Wu \textit{et al.})~\cite{wu2023hidanet} exemplify the dynamic interplay of these two data streams, progressively enhancing the model's representational power and parsing accuracy.

While some models focus primarily on leveraging RGB-D data for improved scene parsing performance, others introduce novel operator designs to extract and capitalize on the complementary information inherent within RGB-D modalities. Methods such as ShapeConv (Cao \textit{et al.})~\cite{cao2021shapeconv} and SGN (Chen \textit{et al.})~\cite{chen2021spatial} propose ingenious techniques to leverage depth-aware convolutions, enabling more nuanced feature extraction and improving object detection in cluttered scenes.

Departing from established paradigms,  our technique employs texture diffusion to refine 1-channel depth representation in higher-level feature spaces. This method aims to bridge the distribution gap between RGB and depth information, substantially reducing information loss during the fusion.  This approach leads to notable advancements in indoor semantic segmentation, salient object detection, and camouflaged object detection, as our empirical evidence will illustrate.

\subsection{RGB-D Salient Object Detection}
The task of Salient Object Detection (SOD) has been revolutionized by the incorporation of depth maps, facilitating a more comprehensive understanding of scene depth and object prominence. Based on their fusion tactics, RGB-D SOD processing methodologies have been organized into three primary frameworks: early fusion~\cite{liu2019salient,zhao2020single,wang2020data,zhang2021uncertainty,chen2021rgb}, intermediate fusion~\cite{liu2021visual,chen2018progressively,liu2021learning,fan2020bbs,fu2021siamese,zhang2021depth,zhang2021bts,liu2021tritransnet}, and late fusion \cite{han2017cnns,wang2019adaptive,ding2019depth}. Zhang et al.~\cite{zhang2021uncertainty} introduced an uncertainty-inspired early fusion approach for RGB-D saliency detection, efficiently merging RGB and depth data to address inherent ambiguities at the input level. Liu et al.~\cite{liu2021learning} devised an intermediate fusion approach using a Selective Mutual Attention and Contrast (SMAC) module to enhance RGB-D saliency detection by leveraging cross-modal interactions and contrast mechanisms. Jin et al.~\cite{jin2022moadnet} utilized asymmetric MobileNetV3 architectures to extract and integrate features from both RGB and depth data effectively, enhancing detection accuracy through structured multi-stage feature fusion. Zhang et al.~\cite{zhang2023multi} develop a multi-prior framework, processing RGB images alongside fine-grained, gradient, and depth priors through individual pipelines to accurately delineate salient objects. Han et al.~\cite{han2017cnns} developed a late fusion strategy that separately processes the high-level representations from both images and depth maps before combining them into a unified saliency map, thereby improving the overall performance of the model. Furthermore, some research has also explored Transformer-based approaches. Liu et al.~\cite{liu2021tritransnet} develop iTransNet, utilizing a Triplet Transformer Embedding Network for enhanced RGB-D salient object detection, emphasizing multi-modal fusion and feature enhancement for improved depth and color integration.

\begin{figure*}[ht!]
  \centering
  \includegraphics[width=\textwidth]{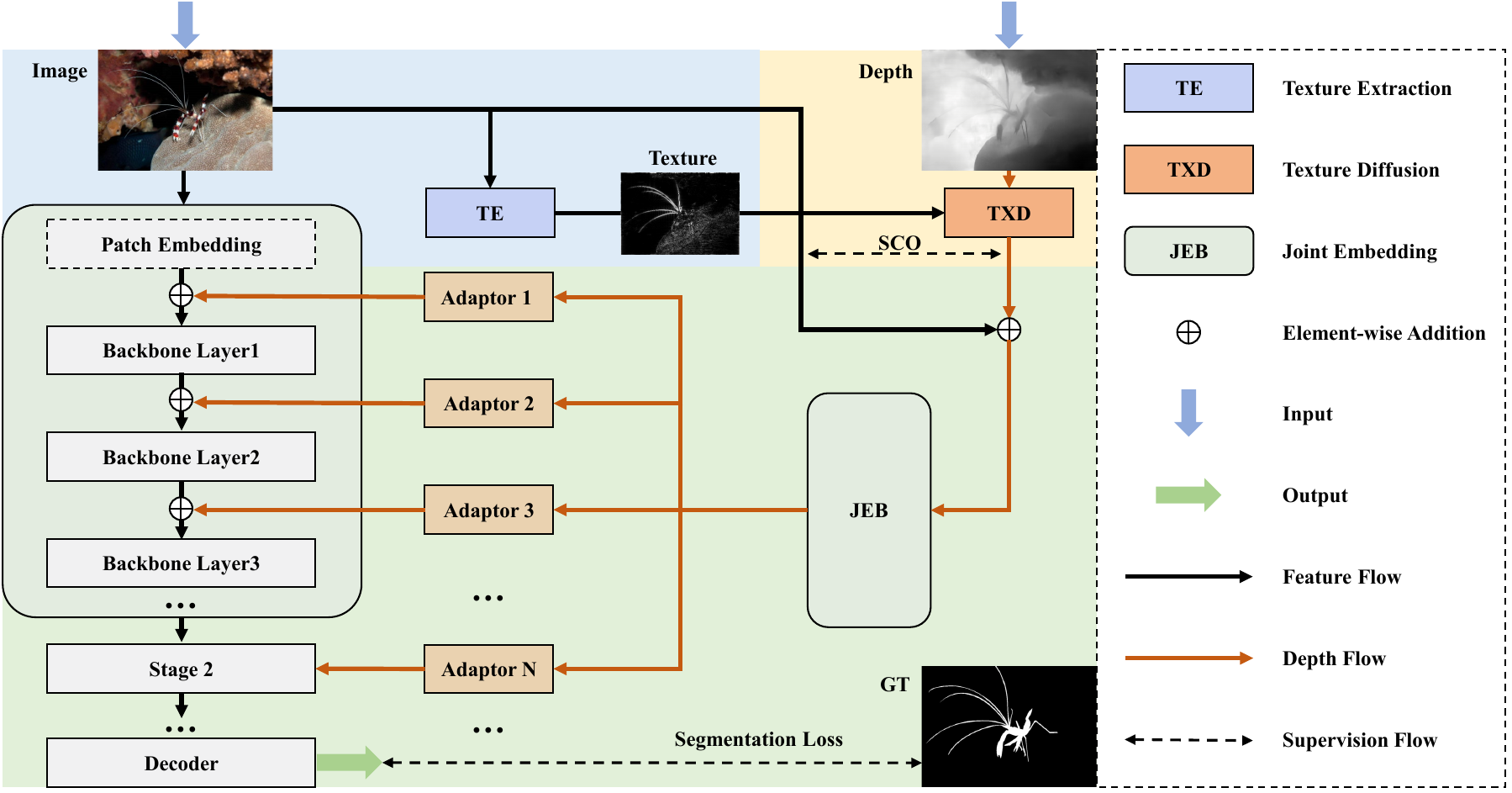} 
  \caption{The architecture of our main framework. It comprises three primary modules: Texture Extraction (TE), which extracts texture features; Texture Diffusion (TXD), which diffuses texture features within depth maps; and Joint Embedding(JEB), which performs joint embedding of the texture-enriched depth and RGB images for improved feature integration.} 
  \label{fig:intro}
\end{figure*}

\subsection{Camouflaged Object Detection}
Historically, camouflaged Object Detection (COD) has capitalized on a range of handcrafted features, such as 3D convexity~\cite{pan2011study}, color~\cite{huerta2007improving}, and edge detection~\cite{siricharoen2010robust}, to identify objects designed to blend into their surroundings. Despite their ingenuity, these approaches often fall short in complex scenarios where the object's concealment is notably challenging. With the advent of extensive COD datasets~\cite{fan2020camouflaged,skurowski2018animal,le2019anabranch,lv2021simultaneously}, there has been a shift towards deep learning models, which offer an enhanced ability to detect camouflaged objects. A pioneering deep learning-based model for COD, inspired by search mechanisms in predatory animals, is SINet~\cite{fan2020camouflaged}, which employs a dual-module strategy for object localization and segmentation. Another noteworthy model is LSR~\cite{he2021enhanced}, which employs a ranking mechanism for detecting and segmenting camouflaged objects, subsequently refining their representation. Efforts to integrate features from interactive learning into graph domains have also been made, exemplified by the two-stage MGL model~\cite{zhai2021mutual}, which enhances the boundary delineation of camouflaged objects. Despite the advancements, these methods sometimes struggle to detect the nuanced clues that differentiate camouflaged objects from their backgrounds. Recent approaches have started to leverage Transformer-based models~\cite{mao2021generative,yang2021uncertainty,liu2023mscaf}, which are adept at capturing the global contextual information necessary for identifying subtle differences in complex scenes. This is particularly relevant to our work, where we incorporate a texture diffusion method that complements these Transformer-based approaches by providing a depth-enhanced textural differentiation for more accurate object detection.

\section{Methodology}
Fig.~\ref{fig:intro} illustrates the overall architecture of our method. For SOD and COD challenges, we employ HitNet~\cite{hu2023high} as the backbone, while for indoor semantic segmentation, DFormer~\cite{yin2023dformer} is utilized, demonstrating our method's adaptability to different backbones.

The architecture is partitioned into three distinct components, each corresponding to a specific colored section in Fig. 2—blue for texture extraction, yellow for texture diffusion, and green for depth integration within segmentation networks. The texture extraction component, primarily consisting of the Texture Extraction (TE) module, is designed to capture and enhance textural features crucial for the subsequent stages. For texture diffusion, the Texture Diffusion (TXD) and Structure Consistency (SC) modules work in tandem to propagate and refine the textural information. Finally, in the depth integration phase, the Joint Embedding(JEB) and Adaptor modules are critical for seamlessly blending depth information into the segmentation networks, ensuring coherent feature synthesis and enhanced segmentation results.

\subsection{Texture Extraction}
Texture and edge details are crucial for capturing important image information and bridging the distribution gap between RGB and depth images. However, directly extracting texture and edge details from the RGB space can often be challenging. Therefore, we extract features from the frequency domain to effectively capture the intricate texture features of objects.

Taking the image \( X \) as input, our method first performs downsampling on \( X \) to obtain a downsampled version \( X_d \). Then, we transform \( X_d \) from the RGB domain to the frequency domain using Fourier Transform (FFT):

\begin{equation}
X_f = \mathcal{F}(X_d)
\end{equation}

where \( X_f \in \mathbb{R}^{3 \times H \times W} \) is the frequency domain representation and \( \mathcal{F} \) denotes FFT. Then we obtain the high-frequency component through a high pass filter, and transform it back to RGB domain to preserve the shift invariance and local consistency of natural images:

\begin{equation}
X_h = \mathcal{F}^{-1}(H(X_f, \alpha))
\end{equation}

where \( H \) denotes the high pass filter and \( \alpha \) is the manually designed threshold which controls the low frequency component to be filtered out. 

\begin{figure}[t]
  \centering
  \includegraphics[width=\columnwidth]{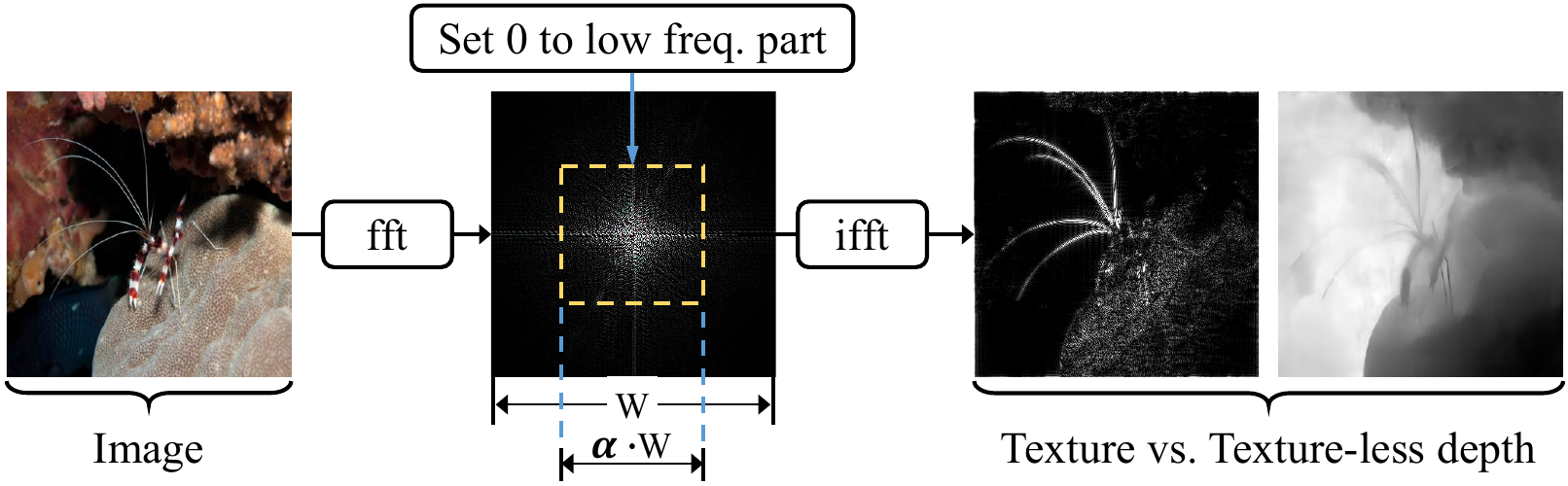} 
  \caption{ The architecture of our Texture Extraction (TE) module.}
  \label{fig:fft}
\end{figure}

As can be seen in Fig.~\ref{fig:fft}, the texture extraction process is exemplified using the case of a marine shrimp. The depth image provides a distinct layout of the scene, which is beneficial for foreground and background separation during the segmentation process. However, it fails to capture the finer details, such as the antennae of the shrimp, which are crucial for a comprehensive scene understanding. Our texture map extraction technique addresses this limitation by recovering these fine details, thereby significantly narrowing the distribution gap between the depth and RGB images. This enhancement ensures that subtle but critical features, like the delicate structures of the shrimp's antennae, are preserved and contribute to the accuracy of the segmentation process.

\subsection{Texture Diffusion}
\subsubsection{Diffusion Process}
Building on the concept of extracting texture and edge details from the frequency domain, as discussed in the previous section, this section focuses on integrating these features into depth maps. Recognizing the inherent texture absence in depth maps, our approach aims to enrich them with the detailed texture features captured from the RGB domain. This process not only enhances the depth maps with essential visual details but also narrows the distribution gap between depth and RGB images, facilitating a more harmonious fusion of the two modalities.

In bridging the gap between depth and RGB images, we recognize the limitations of pixelwise classification, which heavily relies on high-level semantics. To address this, our approach enriches depth maps with detailed texture features derived from the RGB domain, employing reliable low-level semantic cues such as color consistency and pattern smoothness. This strategy enables our model to accurately delineate object boundaries and maintain continuity on small surface structures, especially important in complex scenes where semantic extraction might otherwise lead to misclassification of small structural elements. For instance, rather than solely recognizing an object by its category, our model clusters parts of an object that share similar colors and surface patterns, a process fundamental to our texture diffusion approach. This method, as demonstrated in Fig.~\ref{fig:fft}, facilitates a robust and nuanced integration of texture details within the depth modality. Such an approach not only enhances the depth map with crucial visual details, typically absent but vital for a harmonious fusion, but also ensures the structural integrity of the model across diverse datasets and their inherent scene complexities.

\begin{figure}[t]
  \centering
  \includegraphics[width=\columnwidth]{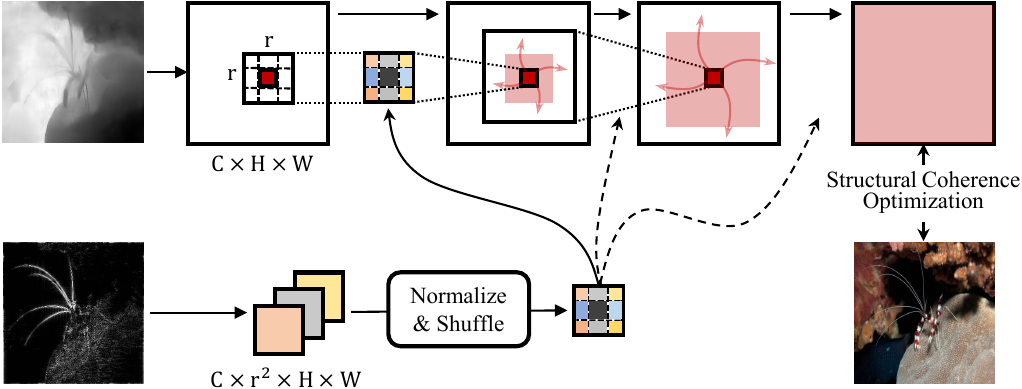} 
  \caption{The architecture of our Texture Diffusion (TXD) module. It utilizes iterative message propagation to diffuse
texture into an enhanced intermediate representation.}
  \label{fig:diffusion}
\end{figure}

In this framework, texture diffusion is conceptualized as a message propagation process within a latent space. Illustrated in Fig.~\ref{fig:diffusion}, the initial step involves using convolutional blocks to convert the depth map \( \mathcal{D} \in \mathbb{R}^{1 \times H \times W} \) into a series of latent features \( \tilde{\mathcal{D}} \in \mathbb{R}^{C \times H \times W} \), where \( C \) denotes the number of latent dims, (e.g., 24). This transformation is crucial for extracting salient features from \( \mathcal{D} \) and enhancing the robustness of message passing against noise.

For each channel, represented as \( \tilde{\mathcal{D}}_i \), we interpret deep pixels as nodes and establish a mechanism for message transfer among proximate nodes. The diffusion weights \( W \in \mathbb{R}^{C \times (r \times r) \times (H \times W)} \) are predicted using convolutional blocks based on the texture feature \( X_h \), where \( r \) is the size of the processing window, (e.g., 7). These weights \( W \) are normalized and shuffled to create distinct kernels \( K_{i,u,v} \in \mathbb{R}^{r \times r} \) for each channel \( i \) and spatial position \( (u, v) \), with the condition \( \sum_{(p,q)} K_{i,u,v}^{p,q} = 1 \). The diffusion unfolds iteratively, with \( K \) employed at each step to update the latent features:

\begin{equation}
\tilde{D}_{i}^{u,v}(t + 1) = \sum_{(p,q) \in N} \tilde{D}_{i}^{p,q}(t) \cdot 
K_{i,u,v}^{p-u+\frac{r}{2},q-v+\frac{r}{2}}
\end{equation}

where \( \tilde{D}_{i}^{(u,v)} \) refers to the feature value at position \( (u, v) \) in channel \( i \), and \( N \) signifies the neighboring locations around \( (u, v) \). The iterative process extends for \(S\) steps to ensure comprehensive message distribution throughout the entire area, enabling each node to gather information from all connected nodes:

\begin{equation}
S = \left\lceil \frac{\max(H,W)}{\left\lfloor \frac{r}{2} \right\rfloor} \right\rceil
\end{equation}

This procedure, despite its iterative nature, remains computationally efficient due to the typically small spatial dimensions of the extracted texture feature (e.g., \( 12 \times 12 \)), complemented by our GPU-optimized implementation which facilitates concurrent message passing for all segments.

\subsubsection{Structural Consistency Optimization}
In the context of depth and RGB image integration, the key challenge is to ensure that the texture-enhanced depth map \(\hat{D}\) aligns structurally with the RGB image \(X\). Structural consistency is crucial, particularly since the texture diffusion process can potentially alter the depth map’s structural integrity. To mitigate this, the Structural Similarity Index (SSIM)~\cite{mac2019digging} is employed as a means to quantify the structural fidelity. This step is vital for seamless depth-RGB fusion and precise semantic segmentation.

Before applying the Structural Consistency Optimization, we employ convolutional blocks to transform the diffusion-enhanced depth image \( \hat{D} \) into a three-channel representation, aligning with the channel dimensions of the RGB image \( X \). Following this transformation, we upscale the modified depth image to match the size of the RGB image \( X \), yielding \( \hat{D}_u \) as a result of the upsampling process:
\begin{equation}
    \hat{D}_u = \text{Upscale}(\text{Conv}(\hat{D}))
\end{equation}
Specifically, to maintain the similarity between texture-enhanced depth map \(\hat{D}_u\) and RGB image \( X \) during optimization, the structural consistency (SC) loss \( L_{\text{SC}} \) is defined as:

\begin{equation}
L_{\text{SC}} = 1 - \text{SSIM}(\hat{D}_u, X)
\end{equation}

Here, the SC loss is weighted by a parameter \( \lambda \), to adjust its influence within the total loss function. Therefore, the total loss function \( L_{\text{total}} \) is expressed as:

\begin{equation}
L_{\text{total}} = \lambda \cdot L_{\text{SC}} + L_{\text{seg}}
\end{equation}

where \( L_{\text{seg}} \) represents the original loss function for the segmentation task. This composite loss function's optimization process ensures that the depth map \( \hat{D} \) is structurally compatible with the RGB image \( X \), which is essential for effective integration and improved segmentation performance.

\subsection{Depth Integration in Segmentation Networks}
\subsubsection{Joint Embedding}
The joint embedding stage is essential for integrating depth information with RGB data. The texture-enhanced depth map \(\hat{D}_u\) is then combined with \( X \) through an element-wise addition:
\begin{equation}
    \quad Z = \hat{D}_u + X.
\end{equation}
The combined representation \( Z \) captures the features from both \( \hat{D}_u \) and \( X \). This representation is processed through an embedding network (e.g., ConvNext) to refine the features for subsequent decoding.

Our custom decoder transforms the output features into a comprehensive joint embedding. The decoder operation is defined as:
\begin{equation}
    F_i = \text{Conv}_i(\text{Upscale}(O_i)), \quad i = 1, 2, 3, 4.
\end{equation}
where \( O_i \) are the outputs from different network stages, and \( \text{Conv}_i \) represents convolutional operations. These feature maps \( F_i \) are concatenated and passed through a fusion convolution layer, represented by \( \mathcal{F} \), to yield the final embedding \( E \):
\begin{equation}
    E = \mathcal{F}(\bigoplus_{i=1}^{4} F_i)
\end{equation}
Here, \( \bigoplus \) denotes the concatenation operation across the feature maps.

\subsubsection{Adaptor}
The Adaptor phase is designed to incorporate depth information into the baseline segmentation network without altering its original structure. This integration enhances the segmentation capabilities while maintaining the integrity of the network's architecture.

The joint embedding \( E \) undergoes an adaptation process to align with the network's layers. This adaptation, denoted by \( A \), which is composed of a stack of convolutions and ReLU activations, adjusts the channel dimensions of \( E \) to produce a series of layer-specific embeddings:
\begin{equation}
    E'_i = A_i(E), \quad  i \in \{1, \ldots, n\}.
\end{equation}
where \( E'_i \) represents the adapted embedding for the \( i \)-th layer of the network, and \( n \) is the total number of layers.

Each \( E'_i \) is resized to match the spatial dimensions of the corresponding layer's input \( x_i \), denoted as \( H_i \times W_i \). The resizing function \( R \) can be formally expressed as:
\begin{equation}
    E''_i = R(E'_i, H_i, W_i), \quad  i \in \{1, \ldots, n\}.
\end{equation}

The resized embeddings \( E''_i \) are then added to the respective inputs of the layers, effectively infusing depth information into the segmentation process:
\begin{equation}
    x'_i = x_i + E''_i, \quad  i \in \{1, \ldots, n\}.
\end{equation}
This approach ensures that the depth information is seamlessly integrated into each layer of the segmentation network, enriching its feature representation and improving segmentation accuracy, all while preserving the network's original configuration and functionality.

\section{Experiments}

\newcolumntype{Y}{>{\centering\arraybackslash}X}
\begin{table*}[t]
\begingroup
\footnotesize
\setlength\tabcolsep{4.2pt}
\renewcommand{\arraystretch}{1}
\begin{center}
\caption{Quantitative comparison on RGB-D SOD datasets with ground truth depth.  $\uparrow$ ($\downarrow$) denotes that the higher (lower) is better.  We use the Mean Absolute Error ($M$), max F-measure ($F_\beta$), S-measure ($S_m$), and max E-measure ($E_\xi$) as evaluation metrics. \textbf{Bold} denotes the best performance.}
\label{tab:sod}
\begin{tabular}{@{} l l | *{4}{c} | *{4}{c} | *{4}{c} | *{4}{c} @{}}
\toprule
\multirow{2}{*}{\ \ \ Public.}  & Dataset &\multicolumn{4}{c}{NLPR} &  \multicolumn{4}{c}{NJUK} & \multicolumn{4}{c}{STERE} & \multicolumn{4}{c}{SIP}\\
\cline{3-6} \cline{7-10} \cline{11-14} \cline{15-18} 

& Metric & 
        \(M \downarrow\) & \(F_\beta \uparrow\) & \(S_m \uparrow\) & \(E_\xi \uparrow\) & 
        \(M \downarrow\) & \(F_\beta \uparrow\) & \(S_m \uparrow\) & \(E_\xi \uparrow\) & 
        \(M \downarrow\) & \(F_\beta \uparrow\) & \(S_m \uparrow\) & \(E_\xi \uparrow\) & 
        \(M \downarrow\) & \(F_\beta \uparrow\) & \(S_m \uparrow\) & \(E_\xi \uparrow\) \\
\hline
\multicolumn{15}{l}{\textbf{Performance of RGB-D Models Trained with GT Depth}} \\
 \ \; TIP$_{21}$ & BIANet~\cite{zhang2021bilateral}
                                   
                                      &  .032&   .888&   .900&   .930 
                                      &  .056&   .878&   .867&   .898
                                      &  .048&   .898&   .895&   .918 
                                      &  .091&   .816&   .802&   .847 \\

 \ \; TIP$_{21}$& HAINet~\cite{li2021hierarchical}
                                       
                                      &  .024&   .920&   .924&   .956 
                                      &  .037&   .924&   .911&   .940
                                      &  .040&   .917&   .907&   .938 
                                      &  .052&   .907&   .879&   .917 \\

 \ \;  TNNLS$_{21}$& D3Net~\cite{fan2020rethinking}
                                       
                                      &  .029&   .904&   .911&   .942 
                                      &  .046&   .909&   .899&   .927
                                      &  .044&   .902&   .906&   .925 
                                      &  .063&   .880&   .860&   .897 \\

 \ \; ECCV$_{22}$ & SPSN~\cite{lee2022spsn}
                                       
                                      &  .023&   .917&   .923&   .956 
                                      &  .032&   .927&   .918&   .949
                                      &  .035&   .909&   .906&   .941 
                                      &  .043&   .910&   .891&   .932 \\

 \ \ \     ICCV$_{23}$ & PopNet~\cite{wu2023source}
                                      &  .019&   .927&   .932&   .963 
                                      &  .030&   .936&   .924&   .952
                                      &  .033&   \textbf{.924}&   .917&   .947 
                                      &  .040&   .923&   .897&  .937 \\ 
\hdashline

  \ \ \     &\textbf{Popnet (HitNet)}
                                      &  .012&   .944&   .944&   .981 
                                      &  .028&   .941&   .928&   .963
                                      &  .030&   .913&   .915&   .952 
                                      &  .030&   .917&   .914&  .954 \\

\rowcolor[RGB]{235,235,250}  

 \ \ \    Ours &\textbf{Ours (HitNet)}
                                      &  \textbf{.010}&   \textbf{.949}&   \textbf{.951}&   \textbf{.984 }
                                      &  \textbf{.023}&   \textbf{.950}&   \textbf{.937}&   \textbf{.971}
                                      &  \textbf{.027}&   .918&   \textbf{.921}&   \textbf{.958 }
                                      &  \textbf{.025}&   \textbf{.928}&   \textbf{.924}&  \textbf{.965} \\

\bottomrule
\end{tabular}
\end{center}
\vspace{-1mm}
\endgroup
\end{table*}

\begin{table*}[t]
\footnotesize
\setlength\tabcolsep{4.2pt}
\renewcommand{\arraystretch}{1.0}
\begin{center}
\caption{Quantitative comparison on COD datasets with source-free depth.}
\label{tab:cod}
\begin{tabular}{@{} l l | *{4}{c} | *{4}{c} | *{4}{c} | *{4}{c} @{}}
\toprule
\multirow{2}{*}{\ \ \ Public.} & Dataset &\multicolumn{4}{c|}{CAMO} &  \multicolumn{4}{c|}{CHAMELEON} & \multicolumn{4}{c|}{COD10K} & \multicolumn{4}{c}{NC4K}\\
\cline{3-6} \cline{7-10} \cline{11-14} \cline{15-18} 
& Metric & 
        \(M \downarrow\) & \(F_\beta \uparrow\) & \(S_m \uparrow\) & \(E_\xi \uparrow\) & 
        \(M \downarrow\) & \(F_\beta \uparrow\) & \(S_m \uparrow\) & \(E_\xi \uparrow\) & 
        \(M \downarrow\) & \(F_\beta \uparrow\) & \(S_m \uparrow\) & \(E_\xi \uparrow\) & 
        \(M \downarrow\) & \(F_\beta \uparrow\) & \(S_m \uparrow\) & \(E_\xi \uparrow\) \\
\hline
\multicolumn{15}{l}{\textbf{Performance of RGB COD Models}} \\
  \quad  CVPR$_{20}$  & SINet~\cite{fan2020camouflaged}
                                       
                                      &  .099&   .762&   .751&   .790
                                      &  .044&   .845&   .868&   .908
                                      &  .051&   .708&   .771&   .832
                                      &  .058&   .804&   .808&   .873 \\

 \quad  CVPR$_{21}$   &  SLSR~\cite{lv2021simultaneously}
                                       
                                      &  .080&   .791&   .787&  .843
                                      &  .030&   .866&   .889&   .938
                                      &  .037&   .756&   .804&   .854
                                      &  .048&   .836&   .839&   .898 \\
 \quad CVPR$_{21}$    & MGL-R~\cite{zhai2021mutual}
                                   
                                      &  .088&   .791&   .775&   .820
                                      &  .031&   .868&   .893&   .932
                                      &  .035&   .767&   .813&   .874
                                      &  .053&   .828&   .832&   .876 \\
 \quad CVPR$_{21} $    &  PFNet~\cite{mei2021camouflaged}
                                       
                                      &  .085&   .793&   .782&   .845
                                      &  .033&   .859&   .882&   .927
                                      &  .040&   .747&   .800&   .880
                                      &  .053&   .820&   .829&   .891 \\
\quad CVPR$_{21}$    & UJSC~\cite{li2021uncertainty}
                                       
                                      &  .072&   .812&   .800&   .861
                                      &  .030&   .874&   .891&   .948
                                      &  .035&   .761&   .808&   .886
                                      &  .047&   .838&   .841&   .900 \\
 \quad IJCAI$_{21}$   & C2FNet~\cite{sun2021context}
                                       
                                      &  .079&   .802&   .796&   .856
                                      &  .032&   .871&   .888&   .936
                                      &  .036&   .764&   .813&   .894
                                      &  .049&   .831&   .838&   .898 \\
 \quad ICCV$_{21}$   & UGTR~\cite{yang2021uncertainty}
                                       
                                      &  .086&   .800&   .783&   .829
                                      &  .031&   .862&   .887&   .926
                                      &  .036&   .769&   .816&   .873
                                      &  .052&   .831&   .839&   .884 \\

  \quad CVPR$_{22}$   & SegMAR~\cite{jia2022segment}
                                      &  .080&   .799&   .794&  .857 
                                      &  .032&   .871&   .887&   .935
                                      &  .039&   .750&   .799&   .876
                                      &  .050&   .828&   .836&   .893 \\

 \quad CVPR$_{22}$  & ZoomNet~\cite{pang2022zoom}
                                      &  .950&   .847&   .894&   .858
                                      &  .033&   .829&   .859&   .915
                                      &  .034&   .771&   .808&   .872
                                      &  .045&   .841&   .843&   .893 \\

\hline

\multicolumn{15}{l}{\textbf{Performance of RGB-D Models Retrained with Source-free Depth}} \\

 \ \;  MM$_{21}$  & CDINet~\cite{zhang2021cross}
                                       
                                      &  .100&   .638&   .732&   .766 
                                      &  .036&   .787&   .879&   .903
                                      &  .044&   .610&   .778&   .821 
                                      &  .067&   .697&   .793&   .830 \\

 \ \; CVPR$_{21}$   &DCF~\cite{ji2021calibrated} 
                                      &  .089&   .724&   .749&   .834
                                      &  .037&   .821&   .850&   .923
                                      &  .040&   .685&   .766&   .864
                                      &  .061&   .765&   .791&   .878 \\

 \ \; ICCV$_{21}$ &CMINet~\cite{zhang2021rgb} 
                                       
                                      &  .087&   .798&   .782&   .827 
                                      &  .032&   .881&   .891&   .930
                                      &  .039&   .768&   .811&   .868 
                                      &  .053&   .832&   .839&   .888 \\

 \ \;  ICCV$_{21}$  &SPNet~\cite{zhou2021specificity}             
                                      &  .083&   .807&   .783&   .831 
                                      &  .033&   .872&   .888&   .930
                                      &  .037&   .776&   .808&   .869 
                                      &  .054&   .828&   .825&   .874 \\
       
 \ \; TIP$_{22}$ &DCMF~\cite{wang2022learning}                             &  .115&   .737&   .728&   .757 
                                      &  .059&   .807&   .830&   .853
                                      &  .063&   .679&   .748&   .776 
                                      &  .077&   .782&   .794&   .820 \\

 \ \;  ECCV$_{22}$  &SPSN~\cite{lee2022spsn} 
                                       
                                      &  .084&   .782&   .773&   .829
                                      &  .032&   .866&   .887&   .932
                                      &  .042&   .727&   .789&   .854 
                                      &  .059&   .803&   .813&   .867 \\

 \ \;   ICCV$_{23}$ & PopNet~\cite{wu2023source} 
                                       
                                      &  .073&   .821&   .806&   .869
                                      &  .022&   .893&   .910&   .962
                                      &  .031&   .789&   .827&   .897
                                      &  .043&   .852&   .852&   .908\\  

\hdashline

 \ \ \      & \textbf{Popnet (HitNet)}  
                                      &  .040& .883  &   .880&   .943 
                                      &  .021& .884  &   .905&   .960
                                      &  .029& .809    & .850&   .927
                                      &  .034& .865  &   .878&  .933 \\ 

\rowcolor[RGB]{235,235,250}
\ \;   Ours & \textbf{Ours (HitNet)} 
                                       
                                      &  \textbf{.029}&   \textbf{.912}&   \textbf{.904}&   \textbf{.964}
                                      & \textbf{.017} &  \textbf{.895} &  \textbf{.912} &  \textbf{.976} 
                                      &  \textbf{.027}&   \textbf{.828}&   \textbf{.861}&   \textbf{.937}
                                      &  \textbf{.030}& \textbf{.883}  & \textbf{.891} &  \textbf{.949} \\  

\bottomrule
\end{tabular}
\end{center}
\vspace{-2mm}
\end{table*}

\subsection{Datasets}

Our experimental evaluation employs datasets across different domains: Salient Object Detection (SOD), Camouflaged Object Detection (COD), and indoor semantic segmentation.

\subsubsection{SOD Datasets}
For SOD datasets, we conduct experiments with the GT depth. We follow the conventional learning protocol~\cite{ji2021calibrated,wu2022robust,zhou2021specificity} and use 700 images from NLPR~\cite{peng2014rgbd} and 1,485 images from NJUK\cite{ju2014depth} for training. The rest are used for testing.
\begin{itemize}
    \item \textbf{NJUK}~\cite{ju2014depth} features 1,985 stereoscopic images tailored for salient object detection.
    \item \textbf{NLPR}~\cite{peng2014rgbd} contains 1,000 stereoscopic images.
    \item \textbf{STERE}~\cite{niu2012leveraging} includes 1,000 stereo images collected from the Internet.
    \item \textbf{SIP}~\cite{fan2020rethinking} offers a high-quality dataset with 929 images focused on salient person detection.
\end{itemize}

\subsubsection{COD Datasets}
For COD datasets, our experiments incorporate the concept of source-free depth, which is inspired by the approaches outlined in the PopNet~\cite{wu2023source}. Following Popnet~\cite{wu2023source}, the state-of-the-art DPT model~\cite{ranftl2021vision} with frozen weights is used as the depth estimation network, which provides us promising source-free depth. We follow the conventional training/testing protocol~\cite{fan2021concealed,fan2020camouflaged,jia2022segment,lv2021simultaneously,pang2022zoom} and use 3,040 images from COD10K~\cite{fan2020camouflaged} and 1,000 images from CAMO~\cite{le2019anabranch} for training. The rest are used for testing.
\begin{itemize}
    \item \textbf{CHAMELEON}~\cite{skurowski2018animal} consists of 76 images that were curated using ``camouflaged animals" as search keywords on Google search.
    \item \textbf{CAMO}~\cite{le2019anabranch} encompasses 1,250 images where each image portrays at least one camouflaged object. Out of these, 1,000 images are allocated for the training set and 250 for the test set. The dataset spans across a multitude of challenging scenarios including variations in object appearance, background clutter, shape complexity, presence of small and multiple objects, occlusions, and potential distractions.
    \item \textbf{COD10K}~\cite{fan2020camouflaged} contains 5,066 images with camouflaged objects. It segregates into 3,040 images for training and 2,026 for testing, further categorized into five super-classes and 69 sub-classes.
    \item \textbf{NC4K}~\cite{lv2021simultaneously} is a testing dataset used in camouflaged object detection, which has a total of 4,121 images. The image scales of these datasets are variable, and there are different levels of camouflage images. In addition, camouflaged objects and salient objects coexist, and art images exist in this benchmark.
\end{itemize}

\subsubsection{Indoor Semantic Segmentation Datasets}
For Indoor Semantic Segmentation Datasets, we conduct experiments with the GT depth. Following common experiment settings~\cite{xie2021segformer,guo2022segnext}, we fine-tune and evaluate the DFormer~\cite{yin2023dformer} on two widely used datasets, NYUDepthv2~\cite{silberman2012indoor} and SUN-RGBD~\cite{song2015sun}. These datasets contain RGB-D samples in various categories and are split into training and testing sets.
\begin{itemize}
    \item \textbf{NYUDepthv2} includes 1,449 RGB-D samples covering 40 categories with a resolution of 480×640, split into 795 for training and 654 for testing.
    \item \textbf{SUN-RGBD} features 10,335 RGB-D images covering 37 categories with a resolution of 530×730, divided into 5,285 for training and 5,050 for testing.
\end{itemize}

\subsection{Evaluation Metrics}

For the evaluation of our models on COD and SOD datasets, we employ the following four widely-used metrics:

\begin{enumerate}
    \item \textbf{Structure-measure} (\(S_m\))~\cite{fan2017structure} evaluates the structural similarity between prediction maps and their corresponding ground truth, closely mirroring human visual perception. It combines object-aware structural similarity (\(s_o\)) and region-aware structural similarity (\(s_r\)), using the formula \(S_m = m \cdot s_o + (1 - m) \cdot s_r\), with \(m\) set to 0.5.

    \item \textbf{F-measure} (\(F_\beta\)) is a metric that balances precision and recall, calculated as \(F_\beta = \frac{(1+\beta^2) \cdot \text{Precision} \cdot \text{Recall}}{\beta^2 \cdot \text{Precision} + \text{Recall}}\), with \(\beta^2\) is set to 0.3 based on previous research. Following previous work [12], we adopt the maximum F-measure as our final evaluation.

    \item \textbf{Enhanced-alignment measure} (\(E_\xi\))~\cite{fan2018enhanced} combines local pixel values with the image-level mean value into a single term. It is calculated as \(E_\xi = \frac{1}{W \times H} \sum_{x=1}^{W} \sum_{y=1}^{H} \theta(\xi)\), where \(\xi\) represents the alignment matrix and \(\theta(\xi)\) denotes the enhanced alignment matrix. Following previous work [12], we adopt the maximum E-measure as our final evaluation.
    
    \item \textbf{Mean Absolute Error} (\(M\)) measures the average absolute difference between the predicted map (\(Prd\)) and ground-truth map (\(G\)), calculated as \(MAE = \frac{1}{W \times H} \sum_{x=1}^{W} \sum_{y=1}^{H} |Prd(x, y) - G(x, y)|\).

\end{enumerate}

For indoor semantic segmentation tasks, we utilize a distinct metric:

\begin{itemize}
    \item \textbf{Mean Intersection over Union} (\(mIoU\)) is the primary metric for evaluating segmentation performance. It averages the IoU across all semantic categories.
\end{itemize}

\begin{figure*}[ht]
    \centering
    \includegraphics[width=\textwidth]{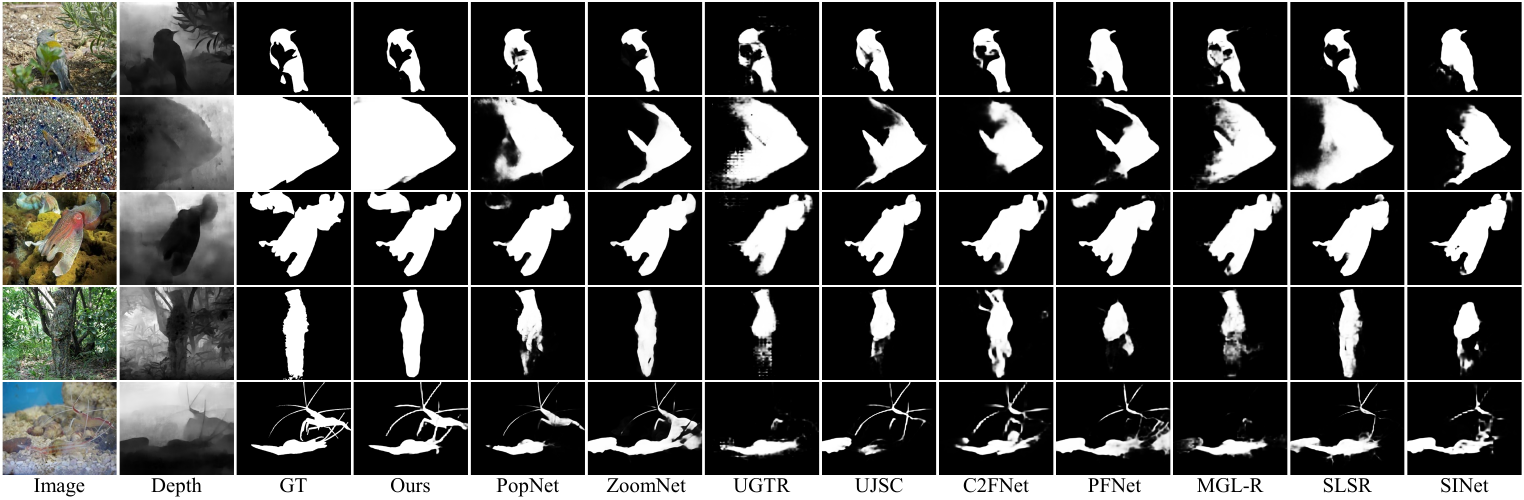}
    \caption{Qualitative comparison of our model against 9 other state-of-the-art methods on the COD datasets.}
    \label{fig:sota_cod}
    \vspace{-1em}
\end{figure*}

\begin{table*}[tp]
\setlength\tabcolsep{4pt}
  \centering
    \scriptsize
    \caption{Quantitative comparison on indoor semantic segmentation datasets with gt depth. $^{\dag}$ indicates our implemented results.}
    \label{tab:semantic}
    \vspace{-2pt}
    \centering
    \renewcommand{\arraystretch}{1}
        \begin{tabular}{@{} l *{1}{p{0.2\textwidth}} *{4}{p{0.15\textwidth}} @{}}
        \toprule
        \textbf{\multirow{2}{*}{Model}}  &\textbf{\multirow{2}{*}{Backbone}}   & \multicolumn{2}{c}{NYUDepthv2} & \multicolumn{2}{c}{SUN-RGBD} \\ \cmidrule(lr){3-4}\cmidrule(lr){5-6}
         & &\textbf{Input size}&  \textbf{mIoU}   &\textbf{Input size}&  \textbf{mIoU} \\
        \midrule\midrule
        ACNet$_{\rm 19*}$~\cite{lee2022spsn}&ResNet-50&$480\times 640$ & 48.3 &$530\times 730$&48.1 \\
        SGNet$_{\rm 20*}$~\cite{chen2021spatial}&ResNet-101&$480\times 640$ & 51.1 &$530\times 730$&48.6\\
        SA-Gate$_{\rm 20*}$~\cite{chen2020bi}&ResNet-101&$480\times 640$ & 52.4 &$530\times 730$&49.4 \\
        CEN$_{\rm 20}$~\cite{wang2020deep}&ResNet-101&$480\times 640$&51.7&$530\times 730$&50.2\\
        CEN$_{\rm 20}$~\cite{wang2020deep}&ResNet-152&$480\times 640$&52.5&$530\times 730$&51.1\\
        ShapeConv$_{\rm 21*}$~\cite{cao2021shapeconv}&ResNext-101&$480\times 640$ & 51.3 &$530\times 730$&48.6 \\ 
        ESANet$_{\rm 21}$~\cite{seichter2021efficient}&ResNet-34&$480\times 640$&50.3&$480\times 640$&48.2\\
        FRNet$_{\rm 22}$~\cite{zhou2022frnet}&ResNet-34&$480\times 640$&53.6&$530\times 730$&51.8\\
        PGDENet$_{\rm 22}$~\cite{zhou2022pgdenet}&ResNet-34&$480\times 640$&53.7&$530\times730$&51.0\\
        EMSANet$_{\rm 22}$~\cite{seichter2022efficient}&ResNet-34&$480\times 640$&51.0&$530\times 730$&48.4\\
        TokenFusion$_{\rm 22*}$~\cite{wang2022multimodal}& MiT-B2&$480\times640$&53.3&$530\times 730$&50.3$^{\dag}$\\
        TokenFusion$_{\rm 22*}$~\cite{wang2022multimodal} &MiT-B3&$480\times640$&54.2&$530\times 730$&51.0$^{\dag}$\\
        MultiMAE$_{\rm 22*}$~\cite{bachmann2022multimae}&ViT-B&$640\times640$&56.0&$640\times 640$&51.1$^{\dag}$\\
        Omnivore$_{\rm 22*}$~\cite{girdhar2022omnivore}&Swin-T&$480\times 640$&49.7&$530\times730$&---\\
        Omnivore$_{\rm 22*}$~\cite{girdhar2022omnivore}&Swin-S&$480\times 640$&52.7&$530\times730$&---\\
        Omnivore$_{\rm 22*}$~\cite{girdhar2022omnivore}&Swin-B&$480\times 640$&54.0&$530\times730$&---\\
        CMX$_{\rm 22*}$~\cite{zhang2023cmx}&MiT-B2&$480\times 640$&54.4&$530\times730$&49.7\\
        CMX$_{\rm 22*}$~\cite{zhang2023cmx}&MiT-B4&$480\times 640$&56.3&$530\times730$&52.1\\
        CMX$_{\rm 22*}$~\cite{zhang2023cmx}&MiT-B5&$480\times 640$&56.9&$530\times730$&52.4\\
        CMNext$_{\rm 23}$~\cite{zhang2023delivering}&MiT-B4&$480\times640$&56.9&$530\times730$&51.9$^{\dag}$\\
        \midrule
        DFormer~\cite{yin2023dformer}&DFormer-L&$480\times 640$ &56.1$^{\dag}$&$530\times730$&51.3$^{\dag}$ \\
        \rowcolor[RGB]{235,235,250}Ours&DFormer-L&$480\times 640$ &\textbf{58.0}&$530\times730$&\textbf{53.2} \\
        \bottomrule
        \end{tabular}
    \hspace{\fill}
    \hspace{\fill}
    \vspace{-10pt}
\end{table*}

\subsection{COD and SOD with texture diffusion}
\subsubsection{Experimental Setup}
We adopt the HitNet~\cite{hu2023high} as backbone for our experiments, pre-trained on the ImageNet dataset. Training images are resized to 384 $\times 384$, with a batch size of 10 for all datasets. The AdamW optimizer is employed with a weight decay set to 0.1. The initial learning rate is set to $5 \times 10^{-4}$. The learning rates for the HitNet backbone and the joint embedding network are adjusted by scaling factors of 0.2 and 0.02, respectively. The learning rate follows a Cosine Annealing schedule. Our experiments are conducted on a pair of NVIDIA GeForce RTX 3090 GPUs.
\subsubsection{Experiments on SOD Datasets}
Table~\ref{tab:sod} provides a quantitative evaluation of our model compared to the state-of-the-art (SOTA) methods on RGB-D SOD datasets. Our method, ``Ours (HitNet)", delineated in the bottom rows, shows an outstanding performance across all datasets. In comparison with PopNet, our model demonstrates an increase in the $F_\beta$ by 2.2\% and in the $E_\xi$ by 2.1\% on the NLPR dataset. On NJUK, our improvements are 1.4\% for $F_\beta$ and 1.9\% for $E_\xi$. For the STERE dataset, the $E_\xi$ is improved by 1.1\%. Lastly, on the SIP dataset, there is an enhancement of 2.7\% in $S_m$ and 2.8\% in $E_\xi$. Notably, even when PopNet employs HitNet as its backbone—referred to as ``PopNet (HitNet)"—our model still outperforms. These results solidify the effectiveness of our method, especially in utilizing sensor depth information to enhance salient object detection accuracy.

\subsubsection{Experiments on COD Datasets}
Table~\ref{tab:cod} illustrates the performance of various COD models on datasets with source-free depth. Our model, delineated in the bottom rows as ``Ours (HitNet)", demonstrates superior performance across all datasets. Specifically, our model outperforms the best-performing previous model, ``PopNet (HitNet)", with a notable margin. In comparison with PopNet, our approach improves the $F_\beta$ metric by 9.1\% and the $E_\xi$ metric by 9.5\% on the CAMO dataset. In the case of the CHAMELEON dataset, our model sees an increase of 0.5\% in $M$ and 1.4\% in $E_\xi$. On the more extensive COD10K and NC4K datasets, our method consistently surpasses the previous state-of-the-art with an improvement of 3.9\% in $F_\beta$ and 4.0\% in $E_\xi$ for COD10K, and 3.1\% in $F_\beta$ and 4.1\% in $E_\xi$ for NC4K. Echoing our success in the SOD experiments, our model not only maintains its superiority when PopNet adopts HitNet as its backbone—now referred to as ``PopNet (HitNet)"—but also shows even higher relative improvements. 

The marked improvements on the COD datasets, particularly on more challenging ones like COD10K and NC4K, underscore a critical aspect of our approach. In complex scenarios like COD, simple fusion techniques that do not account for textural details tend to incur significant information loss. Recognizing this limitation, our method has been specifically designed to address these challenges. These advancements highlight the efficacy of our method, particularly in leveraging source-free depth information to facilitate more accurate camouflaged object detection.

\subsubsection{ Qualitative Comparison}
Fig.~\ref{fig:sota_cod} illustrates the visual comparison among different methods. It can be seen that our model can identify objects on various challenging cases, e.g. occlusion (1st row), similar appearance between background and foreground in terms of color and shapes (2nd, 3rd and 3rd rows), and abundant edge details (5th row). Specifically, our models make satisfactory predictions in three aspects. 1) We can make robust layout perception, i.e. accurately locating the target objects and excluding other distracting regions (1st row). It credits our robust model design and reliable depth integration to alleviate the occlusion problem. 2) We can identify the target objects completely regardless of the background matching (2nd, 3rd and 4th rows). It shows that our texture diffusion enhances the model's perception of object integrity by aggregating low-level features. 3) We can precisely
segment some details of the target objects, e.g. the antennae and legs of the shrimp (5th rows), which shows that our SC loss efficiently preserves the object structure.

\begin{figure*}[ht]
    \centering
    \includegraphics[width=\textwidth]{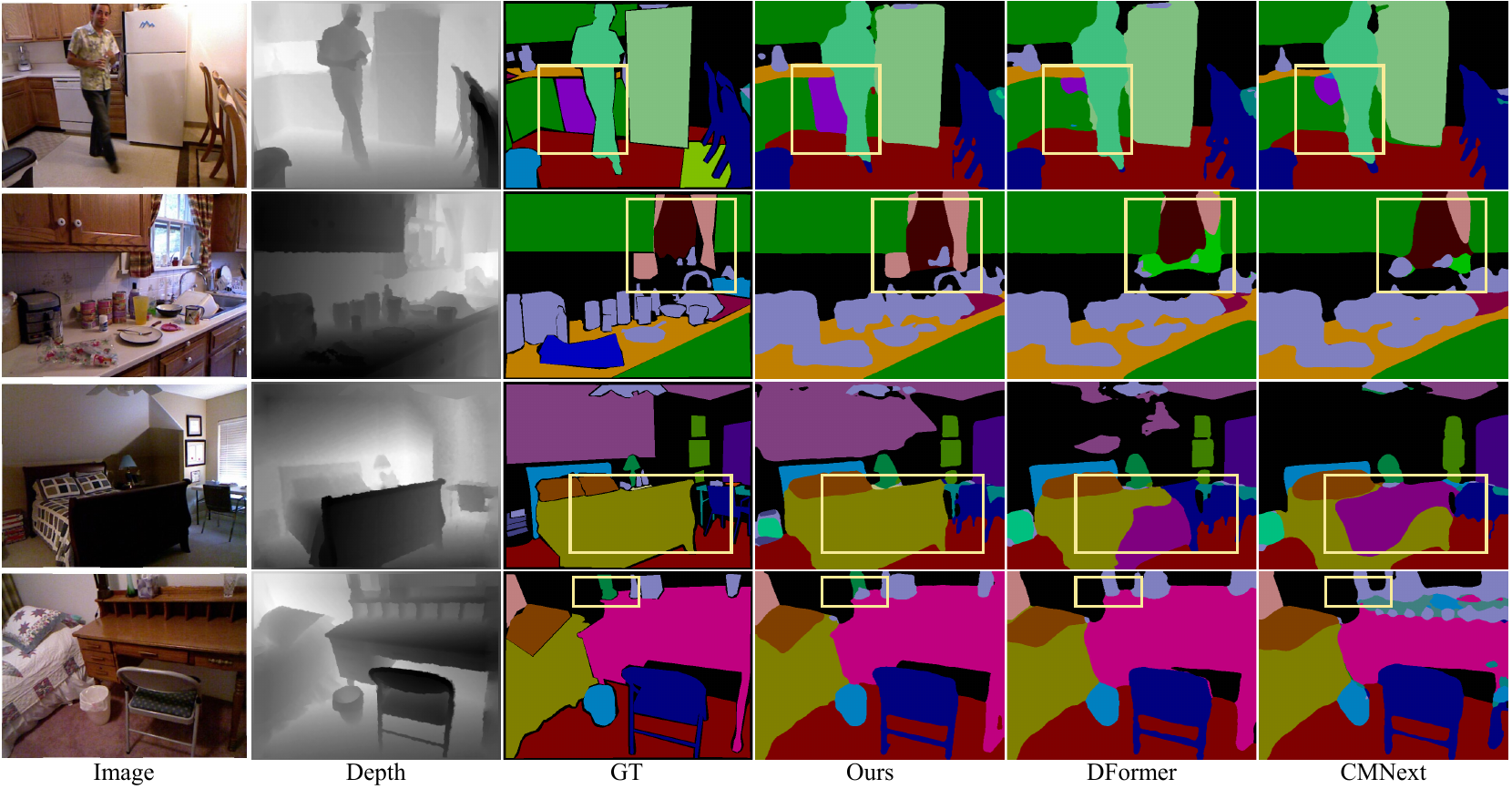}
    \caption{Qualitative comparison of our model against 2 other state-of-the-art methods on the indoor semantic segmentation datasets.}
    \label{fig:sota_indoor}
\end{figure*}

\subsection{Indoor Semantic Segmentation with Texture Diffusion}
\subsubsection{Experimental Setup}
We adopt two common data augmentation strategies: random horizontal flipping and random scaling (from 0.5 to 1.75). For the NYUDepthv2 and SUN-RGBD datasets, the image sizes are respectively set to \(480 \times 640\) and \(530 \times 730\), with a batch size of 6 for both datasets. Cross-entropy loss is utilized as the optimization objective.The weight decay is set to 0.05. The initial learning rates for the NYUDepthv2 and SUN-RGBD datasets are set to \(5 \times 10^{-5}\) and \(3 \times 10^{-5}\), respectively. The learning rate for the joint embedding network is scaled by a factor of 0.2, and a poly decay schedule is employed. Deviating from the approach used in the original DFormer~\cite{yin2023dformer}, we have chosen not to implement any test-time augmentation strategies in both our method and our replication of DFormer.
\subsubsection{Experiments on Indoor Semantic Segmentation Datasets}
The quantitative results of our semantic segmentation experiments are summarized in Table~\ref{tab:semantic}. Our model, which utilizes the DFormer-L backbone~\cite{yin2023dformer}, demonstrates superior performance over the benchmark models on two key datasets. On the NYUDepthv2 dataset, our method achieved a notable mean Intersection over Union (mIoU) of \(58.0\%\), surpassing the DFormer-L's result of \(56.1\%\). This improvement is indicative of the efficacy of our modifications to the original architecture. Turning to the SUN-RGBD dataset, our model's performance is again exemplary, with an mIoU of \(53.2\%\), compared to the \(51.3\%\) achieved by the baseline DFormer-L model. Moreover, the qualitative comparisons between the semantic segmentation results of our method and 2 other state-of-the-art methods in Fig.~\ref{fig:sota_indoor} further demonstrate the advantage of our method. These results affirm the robustness of our approach across different datasets.

 \begin{table*}
\caption{ABLATION STUDY OF THE PROPOSED COMPONENTS}
\label{tab:ablation}
\begin{tabularx}{\textwidth}{@{} l  *{3}{Y} Y| *{3}{Y} Y| *{3}{Y} Y @{}}
\toprule
Settings & \multicolumn{4}{c|}{CAMO} & \multicolumn{4}{c|}{COD10K} & \multicolumn{4}{c}{NC4K} \\
\cmidrule(r){2-5} \cmidrule(lr){6-9} \cmidrule(l){10-13}
&         \(M \downarrow\) & \(F_\beta \uparrow\) & \(S_m \uparrow\) & \(E_\xi \uparrow\) &  
            \(M \downarrow\) & \(F_\beta \uparrow\) & \(S_m \uparrow\) & \(E_\xi \uparrow\) &
            \(M \downarrow\) & \(F_\beta \uparrow\) & \(S_m \uparrow\) & \(E_\xi \uparrow\) \\
\midrule
Baseline &  .039&.878 & 880 & .941 &  .031& .799    & 843&   .921 & .038&  .853 &   .870&  .929 \\
+\(EB\) &  .036&.887 & 885 & .949 &  .029& .810    & 849&   .929 & .035&  .870 &   .879&  .933 \\
+\(JEB\)&  .034&.891 & 889 & .952 &  .029& .814    & 851&   .930 & .034&  .874 &   .883&  .935 \\
+\(JEB\) + \(TXD\)&.030&.905&.901&.962&.028&.823&.856&.935&.031&.881&.887&.948\\
+\(JEB\) + \(TXD\) + \(SC\)&.029&.912&.904&.964&.027&.828&.861&.937&.030&.883&.891&.949\\  
\bottomrule
\end{tabularx}
\vspace{-1em}
\end{table*}

\subsection{Ablation Study}

\begin{figure*}[ht]
    \centering
    \includegraphics[width=\textwidth]{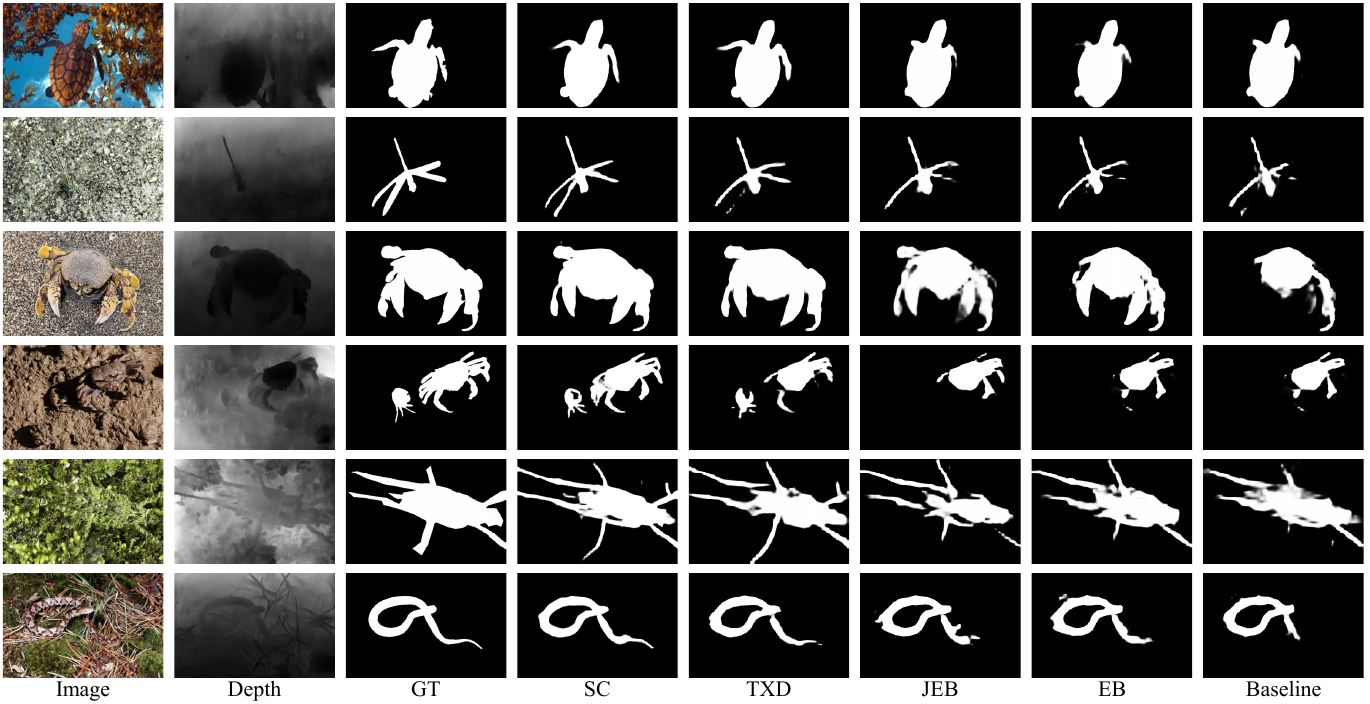}
    \caption{ Visual samples to verify the effectiveness of our proposed components. We show the results of progressively stacking the EB, JEB, TXD, and
SC components on the baseline.}
    \label{fig:ablation}
    \vspace{-2em}
\end{figure*}

In our ablation studies, we aim to demonstrate the effectiveness of the components we propose, the rationale behind the design choices, and the impact of these components specially in the context of COD and SOD datasets.

\subsubsection{Quantitative Component Effectiveness Analysis}
We start by incrementally stacking components on a base HitNet~\cite{hu2023high} structure, to illustrate their contribution to enhancing model performance on COD datasets. Our baseline models are established with HitNet, using only RGB images as inputs. This step-by-step assembly of the model components allows us to track and document the performance improvements throughout the model-building process.
\paragraph{Embedding without Depth Information}
Depth cues are excluded to evaluate the performance contribution from the embedding network alone. By omitting the depth-specific adaptors, the architecture is further simplified. The Embedding features extracted from the RGB images are then adjusted using stage-wise convolutional layers (not layer-wise) to fit the baseline model's dimensions. This variant, referred to as “+EB”(embedding) in Table~\ref{tab:ablation}, helps us measure the effectiveness of the joint embedding network in enhancing the baseline model without the integration of depth data.
\paragraph{Embedding with Depth Information}
 Here we introduce the depth cues into our model. Specifically, we skip the aforementioned texture diffusion process and simply element-wise add the duplicated 3-channel depth maps into the RGB image as input. This simplified multi-modal fusion operation forms the RGBD branch and the whole network is marked as “+JEB”(joint embedding) in Table~\ref{tab:ablation}. 
\paragraph{Diffusion Learning Mechanism}
Given the single channel depth maps, we utilize texture diffusion algorithm to generate the texture-enhanced 3-channel depth maps for the  aforementioned RGBD branch, marked as “+TXD”(texture diffusion) in Table~\ref{tab:ablation}. Our TXD brings significant performance improvements, especially on the MAE metric.
\paragraph{Structural Coherence Optimization}
Finally, we introduce the structure consistency loss to reinforce the structural coherence of the depth map with the RGB image, marked as “+SC” in Table~\ref{tab:ablation}.
SC loss serves as an effective method to align structural details, enhancing the consistency of low-level features across different modalities of data, which promotes a significant performance gain.

\begin{table}[!t]
\centering
\caption{ABLATION STUDY OF DIFFERENT FUSION MANNERS}
\begin{tabularx}{\linewidth}{l *{4}{>{\centering\arraybackslash}X} | *{4}{>{\centering\arraybackslash}X}}
\toprule
\multirow{2}{*}{Method} & \multicolumn{4}{c|}{SIP} & \multicolumn{4}{c}{NC4K} \\
\cmidrule(lr){2-5} \cmidrule(lr){6-9}
 & $M \downarrow$ & $F_\beta \uparrow$ & $S_m \uparrow$ & $E_\xi \uparrow$ & $M \downarrow$ & $F_\beta \uparrow$ & $S_m \uparrow$ & $E_\xi \uparrow$ \\
\midrule
Concatenation & .035 & .908 & .906 & .951 & .030 & .804 & .845 & .928 \\
Hardmard & .032 & .914 & .909 & .953 & .028 & .822 & .855 & .934 \\
Addition & .027 & .928 & .923 & .962 & .027 & .828 & .861 & .937 \\
\bottomrule
\end{tabularx}
\label{tab:fusion}
\end{table}

\subsubsection{Qualitative Component Effectiveness Analysis}
We also provide some visual samples to demonstrate the effectiveness of the aforementioned components. As shown in Fig.~\ref{fig:ablation}, baseline models make many false-negative predictions when encountering complex scenarios because their lack of the depth cues. \textit{EB} and \textit{JEB} can progressively correct some error predictions since they achieve additional high-quality feature representations. Based on them, \textit{TXD} further improves the prediction quality thanks to the reliable enhanced depth maps generated from the texture diffusion. However, the objects predicted by \textit{TXD} are still slightly fragmented, and many details cannot be precisely segmented. The \textit{SC} loss can recover the structural information and precisely segment more details of the target objects.
\subsubsection{Component Design Analysis}
Here we conduct more detailed analyses to verify our design motivations for three components, i.e., JEB, TXD, and SCO. Specifically, we conduct experiments on the baseline and present results on two datasets, i.e., SIP, and COD10K, for quantitative analysis. We also provide some visual samples on NJUK for qualitative analysis.
\paragraph{Analysis of the JEB}
To certify the effectiveness of our feature fusion approach used in the joint embedding, we conducted comparisons between the additive strategy employed in JEB and alternative fusion techniques, such as Hardmard product and concatenation strategy. As demonstrated in Table~\ref{tab:fusion}, our method achieves the best performance when the addition is used.

\begin{table}[!t]
\centering
\caption{ABLATION STUDY OF DIFFERENT ITERATION STEPS}
\begin{tabularx}{\linewidth}{l *{4}{>{\centering\arraybackslash}X} | *{4}{>{\centering\arraybackslash}X}}
\toprule
\multirow{2}{*}{iters} & \multicolumn{4}{c|}{SIP} & \multicolumn{4}{c}{COD10K} \\
\cmidrule(lr){2-5} \cmidrule(lr){6-9}
 & $M \downarrow$ & $F_\beta \uparrow$ & $S_m \uparrow$ & $E_\xi \uparrow$ & $M \downarrow$ & $F_\beta \uparrow$ & $S_m \uparrow$ & $E_\xi \uparrow$ \\
\midrule
1 & .028 & .917 & .915 & .958 & .029 & .819 & .854 & .933 \\
2 & .027 & .921 & .917 & .961 & .028 & .822 & .857 & .935 \\
3 & .025 & .925 & .920 & .965 & .027 & .826 & .860 & .938 \\
4 & .025 & .928 & .924 & .965 & .027 & .828 & .861 & .837 \\
5 & .028 & .919 & .916 & .959 & .027 & .827 & .858 & .936 \\
6 & .028 & .919 & .915 & .959 & .028 & .822 & .855 & .935 \\
\bottomrule
\end{tabularx}
\label{tab:iter}
\end{table}

\begin{figure}[!t]
  \centering
  \includegraphics[width=\columnwidth]{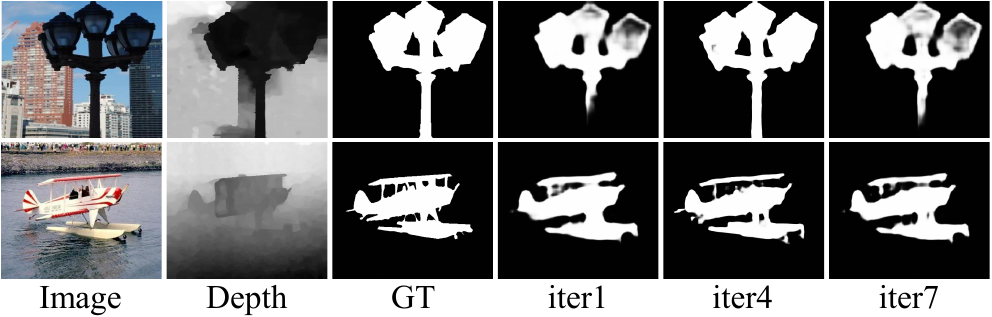} 
  \caption{Visual analysis of the iteration step configuration of the TXD. We show
the visual samples with 1, 3, and 5 iterations based on the HitNet backbones.}
  \label{fig:iter}
  \vspace{-1.5em}
\end{figure}

\paragraph{Analysis of the TXD}
Table~\ref{tab:iter} shows the results of our model with different iteration steps of TXD. We can observe that increasing the iteration step can bring performance improvement, e.g., 1.1\% and 0.9\% improvement of $F_\beta$ on the SIP dataset and COD10K dataset, respectively, when increasing the iteration step from 1 to 4. Additionally, from the visual results in Fig.~\ref{fig:iter}, we can find that using more iterations can effectively exclude incorrect predictions. Such performance improvement begins to saturate when iterating four times. These results indicate that each diffusion step can achieve better results than the previous step until the texture information is diffused across the whole depth image, demonstrating that TXD works following its design motivation.

Moreover, the diffusion kernel size in the TXD process is a key factor affecting feature fusion effectiveness. As detailed in Table~\ref{tab:kernel}, a 7x7 kernel size emerges as the optimal choice. It achieves a delicate balance between efficient information diffusion and the preservation of fine details. This medium-sized kernel ensures robust performance, while both larger and smaller kernels present their unique drawbacks. Larger kernels, although quicker in spreading information, tend to gloss over subtle textural nuances. They may lead to a generalized representation, missing finer details. On the other hand, smaller kernels struggle with limited information reach and increased sensitivity to noise, which leads to an overall suboptimal performance. The 7x7 kernel efficiently propagates texture information throughout the depth image, effectively harnessing the spatial context and preserveing crucial texture details, thus reinforcing the TXD's design principles.

Continuing with our ablation studies, we further examine the texture extraction (TE) mechanism, focusing on the high-pass filter threshold $\alpha$. This parameter is pivotal as it determines the effectiveness of the TE process, which is implemented via a Fourier transform. The experiments revealed that an optimal \(\alpha\) of 0.3 not only enhances texture detail but also suppresses unwanted noise, striking a balance between capturing essential features and avoiding information overload, as shown in Table~\ref{tab:alpha}. This balance is crucial for the model's ability to discern relevant patterns and ignore distractions. In addition to varying \(\alpha\), our analysis extends to contrasting the model's performance with (``w/ TE") and without (``w/o TE") the texture extraction process. The results, as shown in Table~\ref{tab:te}, validate the TE method's effectiveness, where ``w/ TE" significantly enhances the model’s precision, underscoring its crucial role in improving overall performance."

\begin{table}[!t]
\centering
\caption{ABLATION STUDY OF DIFFERENT KERNEL SIZE}
\begin{tabularx}{\linewidth}{l *{4}{>{\centering\arraybackslash}X} | *{4}{>{\centering\arraybackslash}X}}
\toprule
\multirow{2}{*}{kernel} & \multicolumn{4}{c|}{SIP} & \multicolumn{4}{c}{COD10K} \\
\cmidrule(lr){2-5} \cmidrule(lr){6-9}
 & $M \downarrow$ & $F_\beta \uparrow$ & $S_m \uparrow$ & $E_\xi \uparrow$ & $M \downarrow$ & $F_\beta \uparrow$ & $S_m \uparrow$ & $E_\xi \uparrow$ \\
\midrule
3 & .027 & .923 & .919 & .961 & .028 & .823 & .857 & .835 \\
5 & .027 & .923 & .920 & .961 & .028 & .823 & .856 & .836 \\
7 & .025 & .928 & .924 & .965 & .027 & .828 & .861 & .837 \\
9 & .026 & .927 & .921 & .963 & .028 & .824 & .857 & .837 \\
11 & .028 & .921 & .918 & .960 & .028 & .824 & .856 & .837 \\
\bottomrule
\end{tabularx}
\label{tab:kernel}
\end{table}

\paragraph{Analysis of the SCO}
Tuning the loss weight \(\lambda\) is crucial in Structural Coherence Optimization for segmenting complex scenes. The selection of \(\lambda\) subtly weighs structural detail against segmentation consistency. Our results, as presented in Table~X, indicate a notable improvement in performance with a judiciously selected \(\lambda = 0.02\). This enhancement in precision can be attributed to the model's improved ability to conform more closely to true edges while maintaining the smoothness in uniform areas. Such a balance is crucial for the model's effective adaptation to diverse scene complexities.

\begin{table}[!t]
\centering
\caption{ABLATION STUDY OF TEXTURE EXTRACTION (TE)}
\begin{tabularx}{\linewidth}{l *{4}{>{\centering\arraybackslash}X} | *{4}{>{\centering\arraybackslash}X}}
\toprule
\multirow{2}{*}{method} & \multicolumn{4}{c|}{SIP} & \multicolumn{4}{c}{COD10K} \\
\cmidrule(lr){2-5} \cmidrule(lr){6-9}
 & $M \downarrow$ & $F_\beta \uparrow$ & $S_m \uparrow$ & $E_\xi \uparrow$ & $M \downarrow$ & $F_\beta \uparrow$ & $S_m \uparrow$ & $E_\xi \uparrow$ \\
\midrule
w/o TE & .028 & .920 & .917 & .957 & .028 & .821 & .854 & .833 \\
w/ TE & .025 & .928 & .924 & .965 & .027 & .828 & .861 & .837 \\
\bottomrule
\end{tabularx}
\label{tab:te}
\vspace{-1.0em}
\end{table}

\begin{table}[!t]
\centering
\caption{ABLATION STUDY OF DIFFERENT $\alpha$}
\begin{tabularx}{\linewidth}{l *{4}{>{\centering\arraybackslash}X} | *{4}{>{\centering\arraybackslash}X}}
\toprule
\multirow{2}{*}{$\alpha$} & \multicolumn{4}{c|}{SIP} & \multicolumn{4}{c}{COD10K} \\
\cmidrule(lr){2-5} \cmidrule(lr){6-9}
 & $M \downarrow$ & $F_\beta \uparrow$ & $S_m \uparrow$ & $E_\xi \uparrow$ & $M \downarrow$ & $F_\beta \uparrow$ & $S_m \uparrow$ & $E_\xi \uparrow$ \\
\midrule
0.1 & .029 & .919 & .918 & .961 & .028 & .821 & .853 & .832 \\
0.2 & .028 & .922 & .919 & .965 & .027 & .824 & .857 & .834 \\
0.3 & .025 & .928 & .924 & .969 & .027 & .828 & .861 & .837 \\

0.4 & .026 & .927 & .924 & .968 & .027 & .825 & .858 & .834 \\
0.5 & .027 & .925 & .922 & .965 & .027 & .824 & .857 & .833 \\
0.6 & .028 & .923 & .921 & .966 & .028 & .822 & .855 & .832 \\
0.7 & .030 & .917 & .916 & .961 & .028 & .822 & .854 & .831 \\
\bottomrule
\end{tabularx}
\label{tab:alpha}
\vspace{-1.0em}
\end{table}

\begin{table}[!t]
\centering
\caption{ABLATION STUDY OF DIFFERENT $\lambda$}
\begin{tabularx}{\linewidth}{l *{4}{>{\centering\arraybackslash}X} | *{4}{>{\centering\arraybackslash}X}}
\toprule
\multirow{2}{*}{$\lambda$} & \multicolumn{4}{c|}{SIP} & \multicolumn{4}{c}{COD10K} \\
\cmidrule(lr){2-5} \cmidrule(lr){6-9}
 & $M \downarrow$ & $F_\beta \uparrow$ & $S_m \uparrow$ & $E_\xi \uparrow$ & $M \downarrow$ & $F_\beta \uparrow$ & $S_m \uparrow$ & $E_\xi \uparrow$ \\
\midrule
0 & .030 & .921 & .914 & .956 & .028 & .823 & .856 & .835 \\
0.01 & .028 & .922 & .918 & .958 & .027 & .824 & .857 & .837 \\
0.02 & .025 & .928 & .924 & .965 & .027 & .828 & .861 & .837 \\
0.03 & .027 & .926 & .920 & .962 & .027 & .825 & .857 & .836 \\
0.04 & .028 & .924 & .920 & .961 & .028 & .824 & .857 & .835 \\
\bottomrule
\end{tabularx}
\label{tab:lambda}
\vspace{-1.0em}
\end{table}

\section{Conclusion}
In this work, we introduce a Depth-guided Texture Diffusion technique that effectively resolves the fusion issues between depth and vision modalities, enhancing the use of 3D structural information from depth maps for image semantic segmentation. Our method first extracts key texture and edge features from the RGB images, creating a texture map. This texture map is then integrated into the depth map to emphasize structural details crucial for extracting object shapes. By combining this texture-enriched depth with the original RGB data, our approach utilizes depth to improve image segmentation accuracy. Through extensive experiments and ablation studies on commonly used datasets for salient object detection, camouflaged object detection, and indoor semantic segmentation, our method consistently improves baselines and establishes new state-of-the-art results. This research underscores the critical role of texture in depth maps for complex scene interpretation in semantic segmentation, paving the way for future advancements in depth utilization.
\bibliographystyle{IEEEtran}
\bibliography{ieeetran}

\begin{IEEEbiography}[{\includegraphics[width=1in,clip,keepaspectratio]{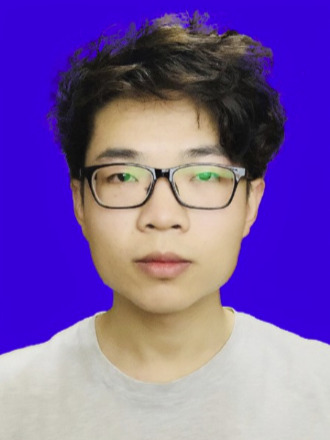}}]{Wei Sun} received the B.S. degree from University of Chinese Academy of Sciences, Beijing, China, in 2020. Since 2020,
 he has been a Ph.D student in the School of Electronic, Electrical and Communication Engineering,
 University of Chinese Academy of Sciences, Beijing, China. His research interests include computer vision and machine learning, specifically for visual
 object segmentation and 3d scene reconstruction.
\end{IEEEbiography}

\begin{IEEEbiography}[{\includegraphics[width=1in,clip,keepaspectratio]{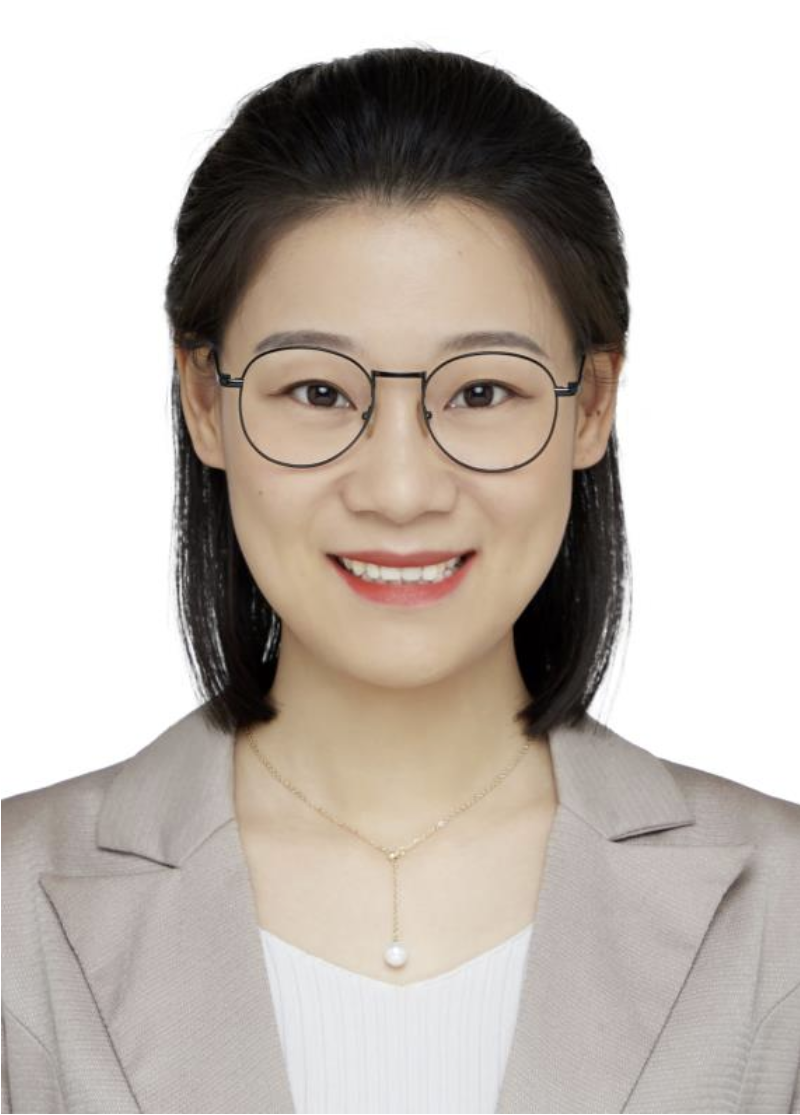}}]{Yuan Li} received the B.S. degree in information management and information system from Dalian Maritime
 University, Dalian, China, in 2010, and the Ph.D degree in logistics engineering in 2022 from the University of Chinese Academy of Sciences, Beijing, China. She has been an associate professor with the University of Chinese Academy of Sciences since 2023. Her research interests include social networks and machine learning.
\end{IEEEbiography}

\begin{IEEEbiography}[{\includegraphics[width=1in,clip,keepaspectratio]{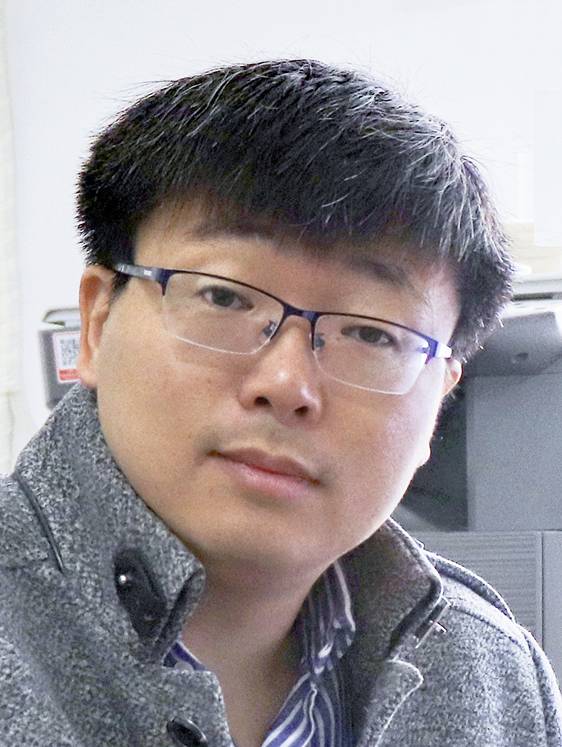}}]{Qixiang Ye} received the B.S. and
 M.S. degrees in mechanical and electrical engineering from Harbin Institute of Technology, China, in 1999 and 2001, respectively, and the Ph.D. degree
 from the Institute of Computing Technology, Chinese Academy of Sciences in 2006. He has been a
 professor with the University of Chinese Academy of Sciences since 2009, and was a visiting assistant
 professor with the Institute of Advanced Computer Studies (UMIACS), University of Maryland, College
 Park until 2013. His research interests include image processing, visual object detection and machine learning. He has published
 more than 100 papers in refereed conferences and journals including IEEE CVPR, ICCV, ECCV, NeurIPS, TNNLS, and TPAMI. He is on the editorial
 boards of IEEE Transactions on Circuit and Systems on Video Technology and IEEE Transactions on Intelligent Transportation Systems.
\end{IEEEbiography}

\begin{IEEEbiography}[{\includegraphics[width=1in,clip,keepaspectratio]{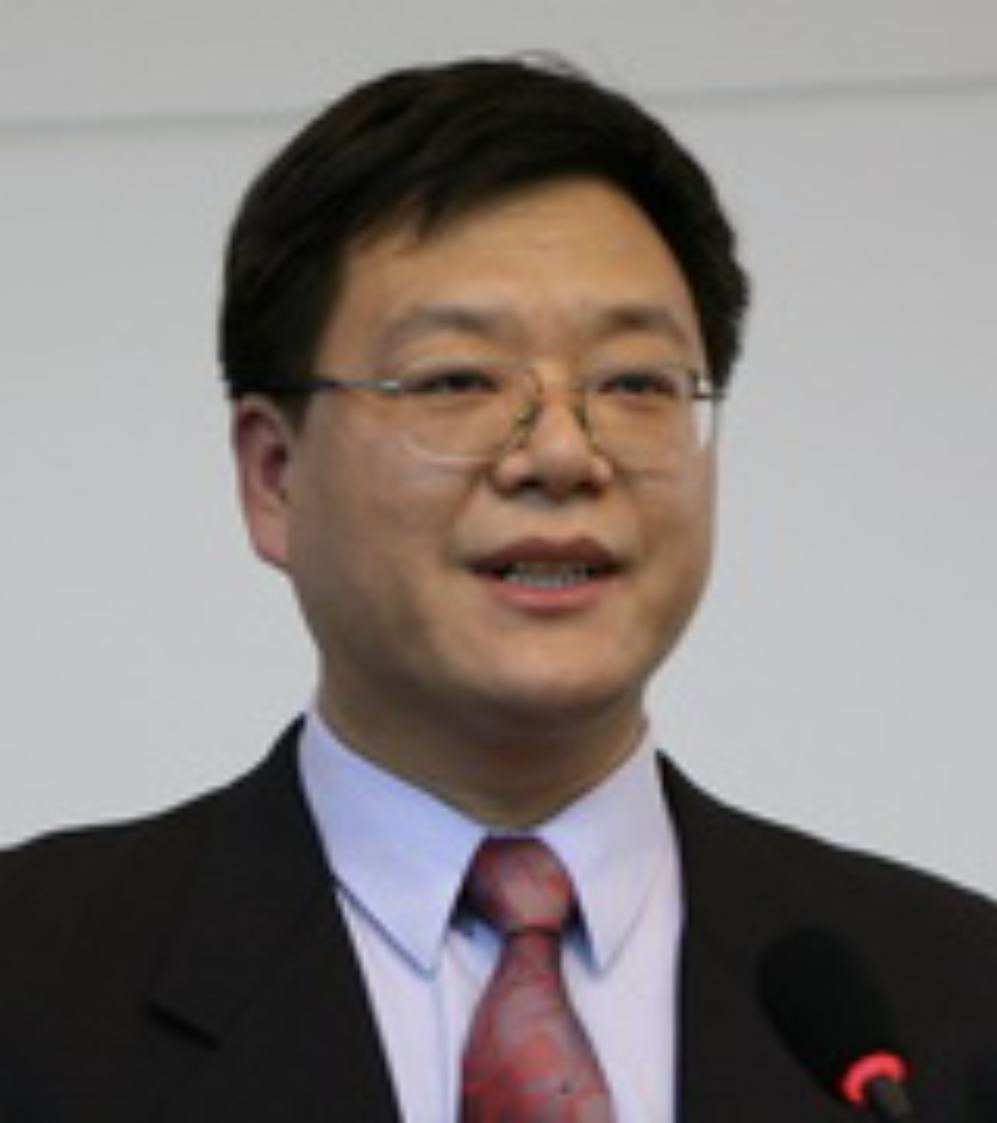}}]{Jianbin Jiao} received the B.S., M.S., and
 Ph.D. degrees from the Harbin Institute of Technology (HIT), China, in 1989, 1992, and 1995, respectively. From 1997 to 2005, he was an associate professor with HIT. Since 2006, he has been a professor with the University of the Chinese Academy of Sciences, Beijing, China. In the research areas about image processing and pattern recognition, he has authored over 50 papers in refereed conferences and journals.
\end{IEEEbiography}

\begin{IEEEbiography}[{\includegraphics[width=1in,clip,keepaspectratio]{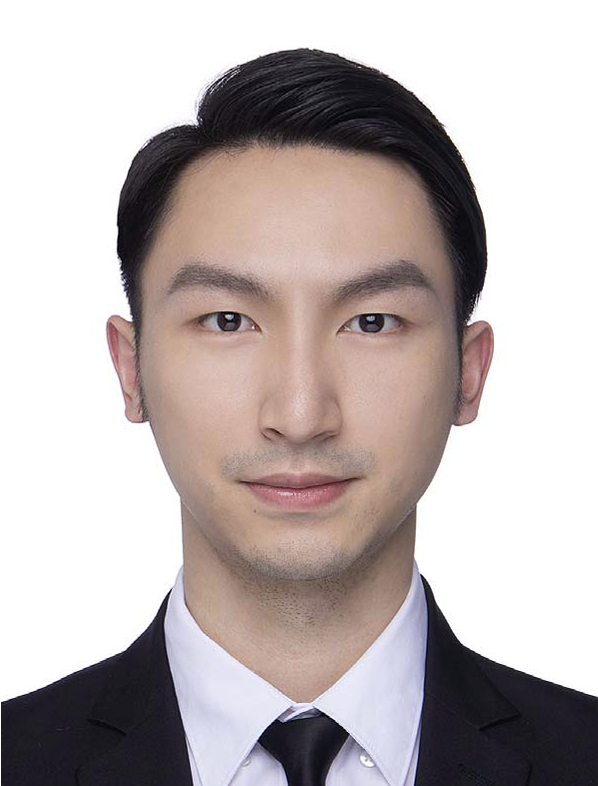}}]{Yanzhao Zhou} received the B.S. degree from Beijing Jiaotong University, China. He received the Ph.D degree
 in signal and information processing in 2022 from the University of Chinese Academy of Sciences, Beijing, China. He has been an associate professor with the University of Chinese Academy of Sciences since 2023. His current research interests include visual perception and multi-modal reasoning.
\end{IEEEbiography}

\end{document}